# A probabilistic patch based image representation using CRF model for image classification

Fariborz Taherkhani

Department of Computer Engineering, Sharif University of Technology, Tehran, Iran

Fariborztaherkhani@gmail.com

**Abstract**

In this paper we proposed an ordered patch based method using Conditional Random Field (CRF) in order to include spatial and geometry properties for image representation to address texture classification, face recognition, and scene classification problems. Typical image classification approaches work without considering spatial causality among distinctive properties of an image for image representation. In this method first, each image is encoded as a sequence of ordered patches, including local properties. Second, the sequence of these ordered patches is modeled as a probabilistic vector by CRF to model spatial relationships of those local properties. And finally, image classification is performed on such probabilistic image representation. Experimental results on several standard image datasets indicate that proposed method outperforms some of existing image classification methods.

## Introduction

Image classification is an interesting and open problem in computer vision. It has many applications in robot navigation, content based image retrieval, organizing photographs [1] and video content analysis [2]. Many methods have been proposed in recent years to classify image categories [3-8]. One of the famous methods is bag of visual words which works based on local patches. Many modified versions of this model have been proposed to enhance the performance of this model in the term of accuracy. Method [9] is one of the efficient methods in this context that extracts a weighted feature vector based on pairwise image matching scheme to select the discriminative features, then visual words are explored using these selected features. One of the drawbacks associated to these methods is that they explore image appearances without considering spatial casualty of discriminative properties of the image [3, 5, 9]. To address this problem new approaches have been proposed [10, 11] to encode spatial information of local properties in the image.

Spatial Pyramid Matching (SPM) is a proper method that uses spatial information of local features [10]. It samples local features from selected grids to include spatial information in feature vectors. It has been shown that SPM works better than bag of visual words model. Bag of spatio-visual words [11] uses spatial context in classification to define a bag of spatio-visual words representation obtained by clustering the visual words' correlogram ensembles. This method uses spherical K-means clustering [12] due to large dimensionality and the sparsity of the proposed spatio-visual descriptors. Method [11] also uses spatial information in the bag of visual words model. This method makes multiple resolution images and obtains local features from each resolution image. To incorporate spatial information, two modalities of horizontal and vertical partitions are performed to divide all resolution images into sub-regions with different scales [11].

In the proposed method our aim is to consider topological and special information using CRF model in image classification problem. An image's spatial context includes those correlated patches which are spatially and topologically adjacent to each other. This fact can be seen from pixel level to semantic level. For example, in the pixel level, typically, the neighbor pixels roughly have similar intensity values and there exists correlation between adjacent pixels of an image. We also have this correlation in semantic level, for example in an image like a beach, typically above of the image is sky, middle of the image is water and the bottom of the image is sand. For most of the global features such as Gist descriptor, we would have correlated extracted values from adjacent patches. To model spatial and topological context in such descriptors, we developed a method that uses CRF with a z-shape chain topology due to existing correlation in both vertical and horizontal directions between adjacent patches. The rest of this paper is organized as follows, in section I, we describe our proposed method and in sections (II) and (III) we discuss about our experimental results and conclusion respectively.

**I) Proposed Approach**

In this section, we describe our proposed model for image classification. This approach includes two parts; First (A), images are encoded in a sequence of ordered patches to extract local properties and second (B), the sequence of these ordered patches is modeled as a probabilistic vector by CRF to model spatial relationships of these local properties.

**I.A) Encoding image as a sequence of ordered patches**

In patch based models, every image is represented as sequence of ordered patches. For example, for a given image X, it is encoded as a sequence of ordered patches such as $X = \{f(x_i) | i = 1,2,3,...n, f(x_i) \in R^m\}$. In this representation $x_i$ is $i^{ith}$ patch in X and n is the number of the patches in the X and $f(x_i)$ is extracted feature vector from patch $x_i$.

In proposed model, local properties and spatial information in the image is encoded in the feature vectors. This model is an appropriate image representation method for face recognition and image classification. First image is represented by local properties. For example street is represented by cars and road in the image along to each other; coast is represented by sky, water and sand from up to down in the image. Second spatial relationship between these local properties is encoded by the order of the patches which include these local properties. For example in face recognition problem, face can be encoded by order of patches which include nose, eyes and mouth for face recognition. Typically there are four different ordering methods (z-shape order [51], Hilbert order [52], row prime order, and row raster order), as shown in Fig.1 to scan patches.

In our model, we observe sequence of the patches in the z-shape order since patches in the image have dependency and correlation in both vertical and horizontal directions and on the other hand in z shape scanning order, patches are observed without losing their dependency in the sequence .

**I.B) CRF model on ordered patches**

For each image in patch based model, the sequence of ordered patches contains local appearance. To include spatial relationships of these local appearance for image representation we use conditional random field (CRF) model.

CRF is a kind of probabilistic graphical model used labeling of sequential data. CRF is a discriminative model that directly estimates the posterior probability and can encode relationships between neighboring observations properly. In CRF, we aim to conditionally model full sequence of labels associate with sequence of the inputs. In ordered patch image representation using CRF model we observe patches sequentially and assign a label to every observed patch in Z- shape scanning order. To model this structure for sequence patches labeling we use linear chain CRF; in other word we aim to model $p(Y|X)$ which is sequence probability of the labels $Y = [y_1, y_2, y_3,...y_n]$ given sequence of ordered patches $X = $

$[x_1, x_2, x_3, \ldots x_n]$ where *n* is number of patches in image X. In this section we formulate image classification problem with our proposed model.

In regular classification problem we assume no dependency between sequences of the training data, in this case the probability of the sequence labels given sequence of input data is formulated as (1)

$$p(Y|X) = \prod_{k=1}^{k=n} p(y_k|x_k) \tag{1}$$

Where, X is sequence of input patches, Y is sequence of labels and $y_k$ is correspond to label of $x_k$ in (1). In order to change the defined term of (1) into the probability close form, we use a partition function *Z(X)* as equation (2)

$$p(Y|X) = \prod_{k=1}^{k=n} exp(p(x_k)y_k)/Z(X) \tag{2}$$

The equation (2) can be written as $p(Y|X) = exp\left(\sum_{k=1}^{k=n} p(x_k)y_k\right)/Z(X)$ in which $p(x_k)y_k$ is the probability of $x_k$ belonging to class $y_k$ and *Z(X)* is a partition function which is equal to the sum of all possible sequences in the numerators, we use this term to normalize the probability to one. However, in context of sequence classification with linear chain we add a term to indicate preference between sequence of labels or adjacent labels because of existing dependency between adjacent sequences of input data. In (3) $Q_{y_k,y_{(k+1)}}$ is the term that expresses preference in the model which is the probability of label $y_k$ followed by label $y_{(k+1)}$ in the model. To consider the whole sequence preference in this model, we successively pair these probabilities overlapping them and then adding each pair as shown in (3)

$$p(Y|X) = exp\left(\sum_{k=1}^{k=n} p(x_k)y_k + \sum_{k=1}^{k=n-1} Q_{y_k,y_{(k+1)}}\right)/Z(X) \tag{3}$$

In (3) partition function Z(X) is summation of all possible label sequences used to normalize summation of probability terms to one. So this partition function is formulated in (4)

$$Z(X) = \sum_{y'_1}\sum_{y'_2}\sum_{y'_3}\ldots\sum_{y'_n} exp\left(\sum_{k=1}^{k=n} p(x_k)y'_k + \sum_{k=1}^{k=n-1} Q_{y'_k,y'_{(k+1)}}\right) \tag{4}$$

(4) represents summation of all possible sequence labels; each $y'_i$ *i={1,2,...n}* in (4) takes all values in set of labels C={$c_1, c_2, \ldots, c_m$} in which *m* is number of all labels.

In order to calculate this term naively, the time complexity would be in exponential order since each of $y'_i$ takes *m* values in the sequence and then with considering the number of all input patches in the sequence which is *n*, the time complexity would be in the order of $O(m^n)$; To address this problem we use forward and backward algorithms using dynamic programing to calculate this partition function. In this approach time complexity of the calculation is reduced from exponential order to polynomial order $O(k.m^2)$; in these methods in each step we record summation result in a table to reuse it for the next step; in forward algorithm we start to sum up the terms (4) from left to right and in backward algorithm we start to sum up the terms (4) from right to left.

The forward and backward algorithm to calculate the partition function in our model is described in part I.B.A and I.B.A respectively.

**I.B.A) Forward Algorithm:**

$FA$:

1) $for\ all\ values\ of\ y'_2\ do$

2) $begin$

3) $\alpha_1(y'_2) \leftarrow \sum_{y'_1} exp\left(p(x_1)y'_1 + Q_{y'_1, y'_2}\right)$

4) $end$

5) $for\ k = 2:n-1\ do$

6) $begin$

7) $for\ all\ values\ of\ y'_{(k+1)}\ do$     (5)

8) $begin$

9) $\alpha_k(y'_{(k+1)}) \leftarrow \sum_{y'_k} exp\left(p(x_k)y'_k + Q_{y'_k, y'_{(k+1)}}\right)\alpha_{(k-1)}(y'_k)$

10) $end$

11) $end$

12) $Z(X) \leftarrow \sum_{y'_n} exp\ (p(x_n)y'_n)\alpha_{(n-1)}(y'_n)$

**I.B.B) Backward Algorithm:**

$BA$:

1) $for\ all\ values\ of\ y'_{(n-1)} do$

2) $begin$

3) $\beta_n(y'_{(n-1)}) \leftarrow \sum_{y'_n} exp\left(p(x_n)y'_n + Q_{y'_{(n-1)},y'_n}\right)$

4) $end$

5) $for\ k = n - 1:2\ do$

6) $begin$

7) $for\ all\ values\ of\ y'_{(k-1)} do$

8) $\quad begin$

9) $\beta_k(y'_{(k-1)}) \leftarrow \sum_{y'_k} exp\left(p(x_k)y'_k + Q_{y'_{(k-1)},y'_k}\right)\beta_{(k+1)}(y'_k)$ (6)

10) $\quad end$

11) $end$

12) $Z(X) \leftarrow \sum_{y'_1} exp\,(p(x_1)y'_1)\beta_2(y'_1)$

For a stable implementation of our algorithm, we work in the logarithm scale. So terms of α and β in *FA* and *BA* are written in logarithm scale as (7) and (8):

$$log\alpha_k(y'_{(k+1)}) \leftarrow log \sum_{(y'_k)} exp\left(p(x_k)y'_k + Q_{y'_k,y'_{(k+1)}} + log\alpha_{(k-1)}(y'_k)\right)$$ (7)

$$log\beta_k(y'_{(k-1)}) \leftarrow log \sum_{(y'_k)} exp\left(p(x_k)y'_k + Q_{y'_{(k-1)},y'_k} + log\beta_{(k+1)}(y'_k)\right)$$ (8)

In I.B.C and I.B.D sections we explain the approaches that we use to estimate the parameters of (3) used in our model. In the other word, we describe the methods used to estimate $p(x_k)y_k$ and $Q_{y_k,y_{(k+1)}}$ parameters respectively.

**I.B.C) Patches distribution using Gaussian Mixture Model**

In this section we utilize probabilistic Gaussian Mixture Model (GMM) to estimate the probability of $x_k$ belonging to class $y_k$, which is the term $p(x_k)y_k$ used in (3).

To calculate this term, for all training images in each class, we encode them as described in I.A section and then we use the standard expectation–maximization (EM) to learn a septate GMM with the parametric set of $\{\pi_i, \mu_i, \Sigma_i, i = 1, \ldots N\}$ in which $\pi_i$ is the mixture weight, $\mu_i$ is the mean vector and $\Sigma_i$ is the (diagonal) covariance matrix of the i [th] Gaussian component .

The algorithm to estimate this parameter is described in more detail in (9).

1) $for\ each\ class\ y_k\ perform$
2) $begin$
3) $training\ set = features\ of\ patches\ in\ class\ y_k$
4) $train\ a\ GMM\ runing\ EM\ for\ this\ set\ of\ features$ (9)
5) $p(x)y_k = \sum_{i=1}^{N} \pi_{ik} G(\mu_{ik}, \Sigma_{ik})$
6) $end$

### I.B.D) Preference term between adjacent labels

Depending on how much patches distributions of two classes are similar to each other, the term of the preference between adjacent labels $y'_k, y'_{(k+1)}$ in our model which is the term of probability $Q_{y'_k, y'_{(k+1)}}$ defined in (3) is determined. The more similarity indicates the higher probability followed by each other. Kullback-Leibler Divergence [53] is usually approximated to measure distance and similarity between two GMMs. Typically, the Monte-Carlo method approximates the Kullback-Leibler Divergence between two Gaussian mixture models $G_a, G_b$ by taking a sufficiently large sampling $S = \{x_1, x_2, x_3, \ldots x_n\}$ which is followed in (10)

$$KL(G_a(x) || G_b(x)) = \int G_a(x) \log \frac{G_a(x)}{G_b(x)} dx \approx \sum_{i=1}^{i=n} \log \frac{G_a(x_i)}{G_b(x_i)} \qquad (10)$$

As much as the distance between two Gaussian mixtures increases then the term of probability $Q_{y'_k, y'_{(k+1)}}$ expressed by each Gaussian mixture decreases. This can be formulated as (11) in which the demonstrator is a normalization term that makes sum of all preference probability from a specific class to other classes equal to one.

$$Q_{y_k, y_{(k+1)}} = \frac{exp\left(-D_{KL}(p(x)y_k z || p(x)y_{(k+1)})\right)}{\sum_{i=1}^{n} exp\left(-D_{KL}(p(x)y_k || p(x)y_i)\right)} \qquad (11)$$

### I.B.D) Probabilistic Image representation for image classification

Suppose that an image belongs to a specific class $c_i$ then all of its patches belongs to $c_i$ as well. Therefore, the marginal probability of single class $c_i$ for all positions in the sequence of the labels $Y = [y_1, y_2, y_3, \ldots y_n]$ is assumed to be higher than this probability for all other classes. Intuitively, we use this pattern of probabilities to create a probabilistic feature vector to represent the image in the feature space.

To represent a given image in CRF probabilistic model, we create a set of probabilistic feature vector $V_{ci}$ in which every element $V_{ci}(k)$, k from position 1 up to position n (number of patches in a given image) expresses $P(X)y_k=c_i$ which is marginal probability of label $c_i$ at position k given the sequence of all patches in the image; we obtain vectors $V_{ci}$ for all labels $\{c_1, c_2, c_3,\ldots c_m\}$ and then sequentially concatenate all of these probabilities vectors $V_{ci}$ together in the feature vector V as an image representative.

For example assume that $X = [x_1, x_2, \ldots, x_n]$ is the sequence of patches of a given image X and $C = [c_1, c_2, \ldots, c_m]$ is the all of the labels; the CRF probabilistic model for image X is represented as V

$$V_{(c_1)} = [p(X)y_1 = c_1, p(X)y_2 = c_1 \ldots, p(X)y_n = c_1]$$
$$V_{(c_2)} = [p(X)y_1 = c_2, p(X)y_2 = c_2 \ldots, p(X)y_n = c_2]$$
$$\vdots \qquad (12)$$
$$V_{(c_m)} = [p(X)y_1 = c_m, p(X)y_2 = c_m \ldots, p(X)y_n = c_m]$$

We then concatenate all of $V_{(c_i)}, i = 1:m$ to create probabilistic feature vector $F = [V_{(c_1)}, V_{(c_2)}, V_{(c_3)}, \ldots, V_{(c_m)}]$ for every image. To calculate marginal probability $p(X)y_k = c_i$ we use alpha and beta table defined in (5) and (6) to calculate it. Alpha table essentially gives summation of term (4) from left to right and beta table gives summation of term (4) from right to left side. The marginal probability of $p(X)y_k = c_i$ is equal to sum over all possible sequence of labels in which $y_k$ is fixed to label $c_i$ at position k, So this marginal probability can be written as

$$p(X)y_k = \underbrace{\sum_{y_1}\sum_{y_2}\cdots\sum_{y_{(k-1)}}}_{\alpha_{(k-1)}(y_k)} y_k \underbrace{\sum_{y_{(k+1)}}\sum_{y_{(k+2)}}\cdots\sum_{y_n}}_{\beta_{(k+1)}(y_k)} exp\left(\sum_{i=1}^{i=n} p(x_i)y_i + \sum_{i=1}^{i=n-1} Q_{y_i,y_{(i+1)}}\right) \Big/ Z(X) \quad (13)$$

The (13) can be simplified as (14); in (14) denominator is a normalization term that makes sum of (14) over all possible label at position k equal to 1.

$$p(X)y_k = \frac{exp\left(p(x_k)y_k + log\alpha_{(k-1)}(y_k) + log\beta_{(k+1)}(y_k)\right)}{\sum_{y'_k} exp\left(p(x_k)y'_k + log\alpha_{(k-1)}(y'_k) + log\beta_{(k+1)}(y'_k)\right)} \tag{14}$$

So the algorithm for image representation in CRF model is described as (15):

*CRF model for image representaion :*

1) *training set*: $X = \{X_1, X_2, \ldots, X_N\}$

2) *for each image* $X_i \in X$ *perform*

3) *divide* $X_i$ *into patches* and *put them* in $Z - shape\ order$: $X_i = [x_1, x_2 \ldots, x_n]$

4) *for each class* $c_j \in C = \{c_1, c_2, \ldots, c_m\}$ *perform*

5) *for each position of label k form* $1:n$ *calculate the marginal probability* as(14):

6) $V_{(c_j)} = [p(X_i)y_k = c_j, p(X_i)y_k = c_j \ldots, p(X_i)y_k = c_j]$

7) *end* (15)

8) *end*

9) $F[X_i] = [V_{(c_1)}, V_{(c_2)}, \ldots V_{(c_m)}]$

10) *end*

Now after representing the image in this form of probabilistic model, we are able to classify them using a classifier.

## II) EXPERIMENTS

In this section, we discuss about the effectiveness and performance of our model for scene classification, face recognition and texture classification problems. We also discuss about sensitivity of our model for tuning parameters including number of Gaussian mixtures, number of patches and order of observing patches in different modes.

### II.A) Experimental setup and discussion

In this model each image is encoded to sequence of some ordered patches. The optimal number of patches for the best performance in this model depends on distribution of the images in the feature space and it should be determined experimentally. The number of the Gaussian components in GMM to estimate the probability distribution of the patches features and class preference probability term should be chosen experimentally as well.

Experimental results in this paper shows that observing patches in Z-shape order mode has higher accuracy rather than other observing modes such as Hilbert, row prime and row (raster) orders as shown in Fig.1. Scanning the patches in z-shape order captures both horizontal and vertical spatial relationships among image patches and also in this mode, patches don't lose their dependency in the sequence for last patch in each row and column. Hilbert order may preserve the locality, however it may bypass either the horizontal or vertical spatial relationships among image patches. Furthermore, a row primer order can better preserve the horizontally spatial relationships among patches, however it may bypass their vertically spatial relationships. For example, as shown in Fig. 1(c), patches 1 and 8 are in vertical contiguous relationship, but the gap in the sequence of ordered patches is very large.

In these experiments, for each patch of an image, we use gist descriptor to describe it.

Gist is a well-known descriptor for scene categorization. This descriptor encodes a set of spatial structure of a scene by using a set of perceptual dimensions including naturalness, openness, roughness, expansion and ruggedness; the degree of these properties can be measured by various techniques such as Fourier transform and PCA. Gist descriptor focuses on the shape of the scene itself and it disregards the local objects in the scene and their relationships. The proposed method by applying CRF model on sequence of ordered patches includes local objects of the image and their spatial relationships information in the Gist descriptor. The Gist as a global descriptor also describes the image as a vector without any interest point.

We compute the 32-dimensional gist global descriptor using 8 orientations, 4 scales for each patch. After image representation using a probabilistic feature vector, we use a RBF kernel of SVM classification to classify images. A one-versus-all method is used to classify multi class problem. We used LIBSVM software package [40] to train and test images.

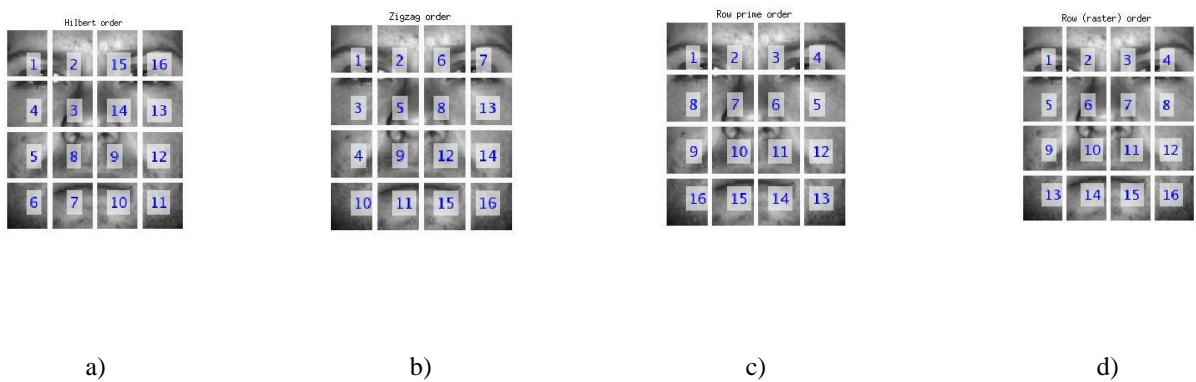

a)          b)          c)          d)

Fig.1: The order of observing patches in different modes. (a) Hilbert (b) Zigzag-shape (c) Row prime order (d) Row (raster) order. The numbers on patches indicates the order of the observed patches in CRF model.

## II.B) Scene classification results

For evaluation of the proposed algorithm, we executed this method over 15 natural scenes[6] which contains 4486 images with 15 classes. This dataset contains 200 to 400 images in each scene category [37]. Experiment runs over 100 randomly selected images per category for training and the rest are used for testing. To obtain the best performance, the experimental results in Fig.2 show that the number of Gaussian components and the number of patches for image representation in this model should be fixed to 4 and 9 respectively. Fig.2 shows the performance of the algorithm in the terms of the accuracy for 15 natural set with different number of patches and Gaussian components in GMM model.

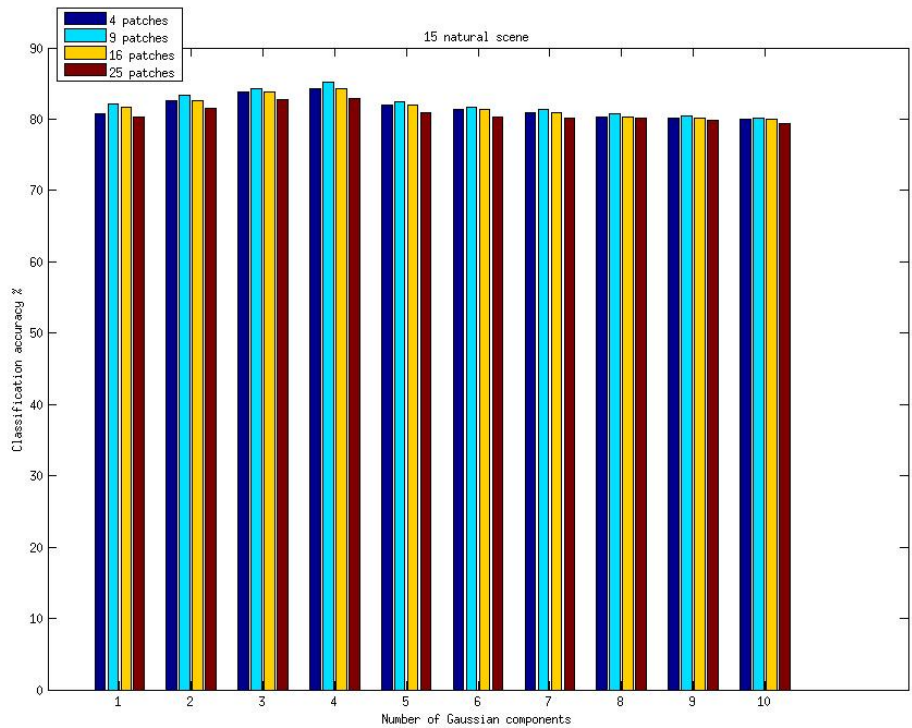

Fig.2: Accuracy of the algorithm corresponding to different patches and Gaussian components in GMM for 15 natural scene dataset.

Fig.2 shows the performance of the algorithm in the terms of the accuracy for 15 natural set with different number of patches and Gaussian components in GMM model. For each number of ordered patches with 4,9,16 and 25, we calculated the accuracy of our algorithm with different Gaussian components from 1 to 10.

Fig.3 shows the per class accuracy for 15 natural scene dataset. The proposed algorithm performance for outdoor classes such as CALsuburb, MITcoast, MITforest and MITcountry is higher than indoor classes such as Kitchen, Paroffice and bedroom.

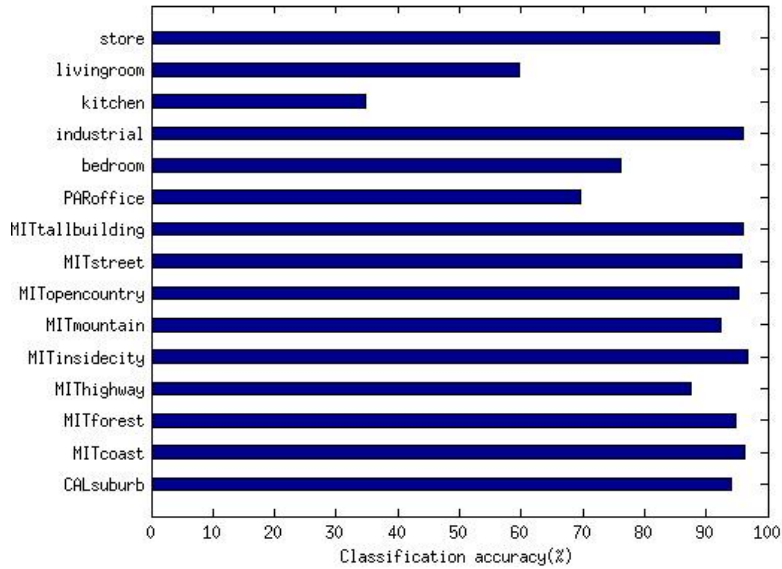

Fig.3: per class accuracy rate for 15 natural scene dataset

**II.C) UIUC texture classification results**

The UIUC texture dataset has 1000 images with 25 texture classes, and each class consists of 40 images with size of 640*480 pixels. The textures in this dataset have discriminative viewpoint and scale variations with variant illumination and contrast. We perform our experiment with 20 randomly selected images in each class for training and the rest for test. The results show that our approach works better than some other exiting methods except for WaveLBPS + GMM [48].

Fig.4 shows the performance of the algorithm in the terms of the accuracy for UIUC texture dataset with different number of patches and Gaussian components in GMM model. Fig.4 shows that the number of the patches and Gaussian components should be fixed to 16 and 7 respectively for obtaining of the best performance on UIUC texture dataset.

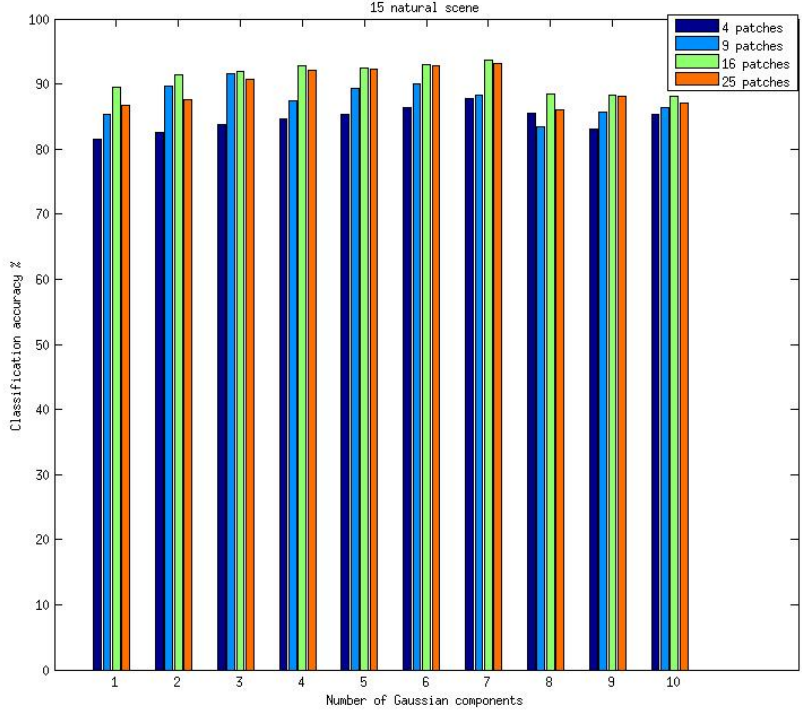

Fig.4: Accuracy of the algorithm corresponding to different patches and Gaussian components in GMM for UIUC texture dataset.

Table (1) indicates the average classification accuracy of proposed approach and some other methods over the 15 scene and UIUC texture datasets. The results show that our method is better than some other exiting methods.

| Methods | Experimental dataset | |
|---|---|---|
| | UIUC texture | 15 natural scene |
| WaveLBPS + GMM  [48] | 94.3% | 80.6% |
| Db4_LBPS + GMM  [48] | 93.5% | 79.8% |
| WaveLBPM + GMM [48] | 93.6% | 81.3% |
| Db4_LBPM + GMM  [48] | 93.4% | 81.5% |
| Proposed Method | 93.75% | 85.18% |

Table (1): Classification accuracy on UIUC texture and 15 natural scene datasets.

## II.D) Face recognition results

To evaluate our algorithm for face recognition problem, we run our algorithm on three standard face recognition datasets including Yale, Olivetti-Oracle Research Lab (ORL) [49], and extended Yale B [50]. For face recognition, Table (2) shows the comparison recognition accuracy. The first five rows show the classifier performance using the method of Principal Component Analysis (PCA), Linear Discriminant Analysis (LDA), Locality Preserving Projections (LPP), Implicit Elastic Matching (IEM) [46] and RML [47] respectively. Experimental results show our proposed method outperforms the other approaches on three aforementioned databases. The best performance for Yale , Extended Yale and ORL dataset are obtained once we set parameters of patches numbers to 3,2,2 and Gaussian components to 2,6,3 respectively.

| Method | Experimental dataset | | |
|---|---|---|---|
| | Yale | ORL | Ext.Yale B |
| Wright(PCA)[46] | 80% | 88.1% | 65.4% |
| Wright(LDA)[46] | 87.3% | 93.9% | 81.3% |
| Wright(LPP)[46] | 88.6% | 88.59% | 86.4% |
| Wright(IEM)[46] | - | 96.5% | 91.4% |
| RML [47] | 85.6% | 90% | - |
| Proposed Method | 94.6% | 97.5% | 98.62% |

Table (2): Face recognition accuracy rate for different method on Yale, ORL and Ext.Yale B datasets.

In order to analyze the sensitivity of our proposed algorithm to the order of the patches observed we perform experiment on the sequence of the patches in four different methods including (row raster order, row prime order, Hilbert order [52] and z-shape order [51]) as shown in Fig. 1. Recognition accuracy on three well-known face recognition databases (Yale, extended Yale B and ORL) are indicated in Table (3). Experimental results show that different ordering modes have different recognition rates. The recognition rate of z-shape scanning mode on three aforementioned datasets works closely better than other modes. Because scanning the patches in the z-shape mode encodes both horizontal and vertical spatial casualty information among image patches. Hilbert order has potential of preserving the locality of the image. However, it may bypass either the horizontal or vertical spatial casualty information among image patches. And also row primer scanning mode can only preserve spatial causality information horizontally.

| Dataset | Row(raster) | Row( prime) | Hilbert | Z-shape |
|---------|-------------|-------------|---------|---------|
| Yale | 93.30% | 93.30% | 91.3% | 94.6% |
| Ext.Yale | 98.42% | 98.56% | 98.3% | 98.62% |
| ORL | 96.5% | 96.50% | 92.50% | 97.5% |

Table (3): Face recognition rate for different ordering methods on Yale, Ext.Yale and ORL datasets.

Fig.5 shows some of images from 15 natural scene, UIUC texture datasets and aforementioned face recognition datasets.

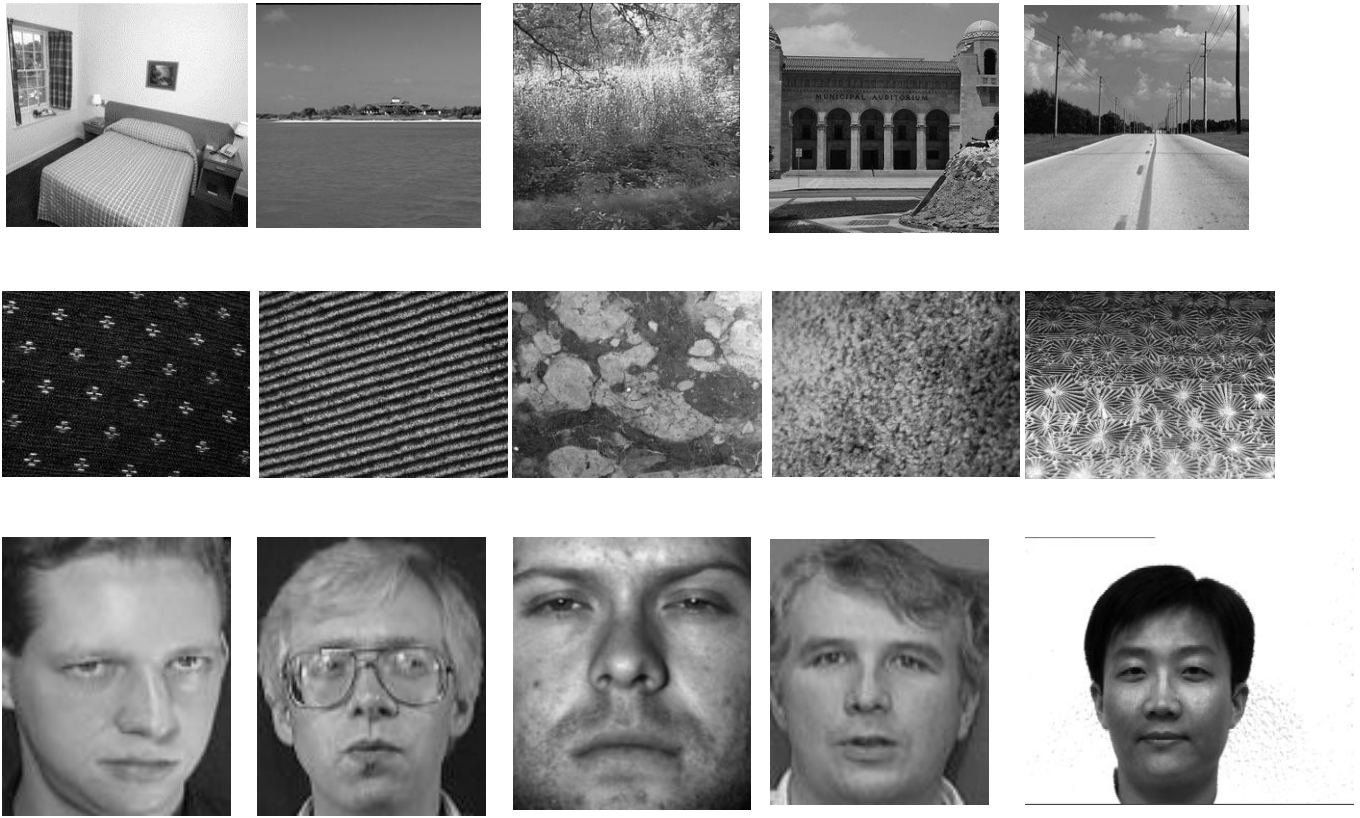

Fig.5 Samples of images in datasets used in our experiments: 15 Natural scene (first row), UIUC texture dataset (second row) and some of samples from aforementioned face recognition datasets (third row). Each sample belongs to one image class in the defined dataset.

## III) Conclusion

We proposed a probabilistic patch based image representation for image classification using CRF model with high classification accuracy. The proposed method encodes local properties and their spatial relationship in the images to represent them in the feature space. The evaluation results on standard face recognition, scene and texture datasets showed superiority of this method in comparison to some of existing methods.